\documentclass[11pt, a4paper, logo, onecolumn]{googledeepmind}

\usepackage[authoryear,  round]{natbib}
\usepackage{subfigure}
\usepackage{multirow}
\usepackage[subtle]{savetrees}

\title{Griffin: Mixing Gated Linear Recurrences with Local Attention for Efficient Language Models}

\correspondingauthor{ \{sohamde, slsmith\}@google.com}

\reportnumber{} %

\usepackage{geometry}
\usepackage[parfill]{parskip}
\usepackage{amsmath,amsfonts,amsthm}
\usepackage{booktabs}
\usepackage[authoryear]{natbib}
\usepackage{graphicx}
\usepackage{xcolor}
\usepackage[capitalise]{cleveref}
\usepackage{bm}
\usepackage{colortbl}

\usepackage{listings}
\usepackage{color}

\definecolor{dkgreen}{rgb}{0,0.6,0}
\definecolor{gray}{rgb}{0.5,0.5,0.5}
\definecolor{mauve}{rgb}{0.58,0,0.82}

\definecolor{tablerowcolor}{RGB}{240, 245, 255}
\definecolor{tableemphcolor}{RGB}{0, 0, 200}

\DeclareMathOperator{\sigmoid}{sigmoid}
\DeclareMathOperator{\softplus}{softplus}

\theoremstyle{definition}

\newcommand{\subblock}[0]{{block}}
\newcommand{\subblocks}[0]{{blocks}}

\author[* 1]{Soham De}
\author[* 1]{Samuel L. Smith}
\author[*1]{Anushan Fernando} \author[*1]{Aleksandar Botev} \author[*1]{George Cristian-Muraru} \author[2]{Albert Gu} \author[]{Ruba Haroun} \author[]{Leonard Berrada} \author[]{Yutian Chen} \author[]{Srivatsan Srinivasan} \author[]{Guillaume Desjardins}
\author[]{Arnaud Doucet}
\author[]{David Budden}
\author[]{Yee Whye Teh} \author[]{Razvan Pascanu} \author[]{Nando De Freitas} \author[]{Caglar Gulcehre}

\affil[*]{Equal contributions}
\affil[1]{Google DeepMind}
\affil[2]{Work done while at Google DeepMind}

\begin{abstract}
Recurrent neural networks (RNNs) have fast inference and scale efficiently on long sequences, but they are difficult to train and hard to scale. We propose Hawk, an RNN with gated linear recurrences, and Griffin, a hybrid model that mixes gated linear recurrences with local attention. Hawk exceeds the reported performance of Mamba on downstream tasks, while Griffin matches the performance of Llama-2 despite being trained on over 6 times fewer tokens. We also show that Griffin can extrapolate on sequences significantly longer than those seen during training. Our models match the hardware efficiency of Transformers during training, and during inference they have lower latency and significantly higher throughput. We scale Griffin up to 14B parameters, and explain how to shard our models for efficient distributed training.

\end{abstract}

\begin{document}

\maketitle

\section{Introduction}

Recurrent neural networks (RNNs) played a central role in the early days of deep learning and NLP research \citep{elman1990finding, Siegelmann1991, hochreiter1997long, Mikolov2010, bahdanau2014neural, sutskever2014sequence}, and achieved practical success in many applications, including Google's first end to end machine translation system \citep{wu2016google}. 
However in recent years, both deep learning and NLP have been dominated by the Transformer architecture \citep{vaswani2017attention}, which interleaves multi-layer perceptrons (MLPs) and multi-head attention (MHA). 
Transformers achieve better performance than RNNs in practice and are also very efficient at utilizing modern hardware  \citep{kaplan2020scaling}.
Transformer-based large language models trained on massive datasets collected from the web have achieved remarkable success \citep{brown2020language,rae2021scaling,  hoffmann2022training, touvron2023llama, achiam2023gpt, team2023gemini}.

Despite their successes, Transformers are difficult to scale efficiently to long sequences due to the quadratic complexity of global attention.
Additionally, the linear growth of the Key-Value (KV) cache with the sequence length makes Transformers slow during inference.
Although Multi-Query Attention (MQA) \citep{shazeer2019fast} partially mitigates this issue by reducing the cache size by a constant factor, the cache still grows linearly in sequence length.
Recurrent language models present a compelling alternative as they compress the entire sequence into a fixed-sized hidden state which is updated iteratively. 
However to replace Transformers, new RNN models must demonstrate not only comparable performance at scale but also achieve similar hardware efficiency \citep{gu2021efficiently,mehta2022long, smith2022simplified, orvieto2023resurrecting, dao2022hungry, poli2023hyena, gu2023mamba}.

\begin{figure}[t]
\centering
\subfigure[Scaling curve during training]
{\includegraphics{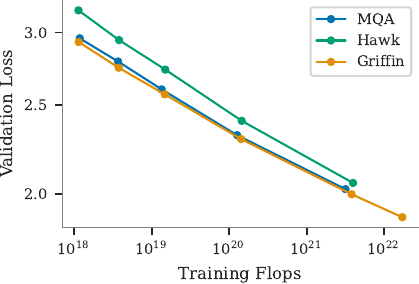}\label{fig:intro_a}}
\subfigure[Maximum throughput at 1B parameter scale.]
{\includegraphics{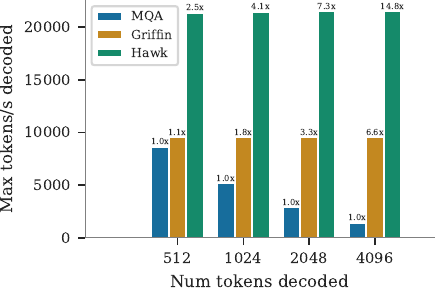}\label{fig:intro_b}}
\caption{a) Hawk, Griffin and our MQA Transformer baseline all show power law scaling between held-out loss and training FLOPs, with Griffin achieving the lowest held-out loss at all FLOPs budgets. The largest Griffin model shown has 14B parameters. b) Hawk and Griffin achieve significantly higher throughput than our MQA Transformer, especially when the length of the sample increases.}
\label{fig:intro}
\end{figure}

In this work, we propose the RG-LRU layer, a novel gated linear recurrent layer, around which we design a new recurrent block to replace MQA. 
We build two new models using this recurrent block: \textbf{Hawk}, a model which interleaves MLPs with recurrent blocks, and \textbf{Griffin}, a hybrid model which interleaves MLPs with a mixture of recurrent blocks and local attention \citep{beltagy2020longformer}. We show that:

\begin{enumerate}
    \item Hawk and Griffin exhibit power law scaling between held-out loss and training FLOPs, up to and beyond 7B parameters (Figure \ref{fig:intro_a}), as previously observed for Transformers \citep{kaplan2020scaling}. \vspace{2mm}
    \item Griffin achieves slightly lower held-out loss than strong Transformer baselines at all model scales. \vspace{2mm}
    \item We overtrain Hawk and Griffin on 300B tokens at a range of model scales. Hawk-3B exceeds the reported performance of Mamba-3B \citep{gu2023mamba} on downstream tasks, despite being trained on half as many tokens. Griffin-7B and Griffin-14B match the performance of Llama-2 \citep{touvron2023llama} despite being trained on roughly 7 times fewer tokens (Section \ref{sec:downstream_evals}). \vspace{2mm}
    \item Both Hawk and Griffin achieve comparable training efficiency to Transformers on TPU-v3. Since diagonal RNN layers are memory bound, we achieve this with a kernel for the RG-LRU layer, implemented in Pallas \citep{jax2018github}, that minimizes memory transfers (Section \ref{sec:training_efficiency}). \vspace{2mm}
    \item During inference, both Hawk and Griffin achieve significantly higher throughput than MQA Transformers (Figure \ref{fig:intro_b}), and they achieve lower latency when sampling long sequences (Section \ref{sec:inference}). \vspace{2mm}
    \item Griffin performs better than Transformers when evaluated on sequences longer than those seen during training, and can also efficiently learn copying and retrieval tasks from training data (Section \ref{sec:long_seq}). However, Hawk and Griffin perform less well than Transformers when we evaluate pre-trained models on copying and exact-retrieval tasks without fine-tuning.
\end{enumerate}

\section{Model Architecture}
\label{sec:model}

All our models contain the following components: (i) \emph{a residual block}, (ii) \emph{an MLP \subblock{}}, and (iii) \emph{a temporal-mixing \subblock{}}. 
While (i) and (ii) are the same across all models, we consider three temporal mixing \subblocks{}: \emph{global Multi-Query Attention} (MQA), \emph{local (sliding-window) MQA} and our proposed \emph{recurrent \subblock{}}. 
As part of the recurrent \subblock{} we use the Real-Gated Linear Recurrent Unit (RG-LRU) -- a novel recurrent layer  inspired by the Linear Recurrent Unit~\citep{orvieto2023resurrecting}.

The residual block, as shown in Figure~\ref{fig:model_sketch_1}(a), defines the global structure of our models and is inspired by pre-norm Transformers \citep{xiong2020layer}. 
After embedding the input sequence we pass it through $N$ such blocks ($N$ denoting the model depth), and then we apply RMSNorm \citep{zhang2019root} to produce the final activations.
To compute the token probabilities we apply a final linear layer followed by a softmax.
The weights of this layer are shared with the input embedding layer.

\subsection{Residual block}
\label{sec:residual_pattern}
\begin{figure}[t]
\centering
\includegraphics[width=0.91\linewidth]{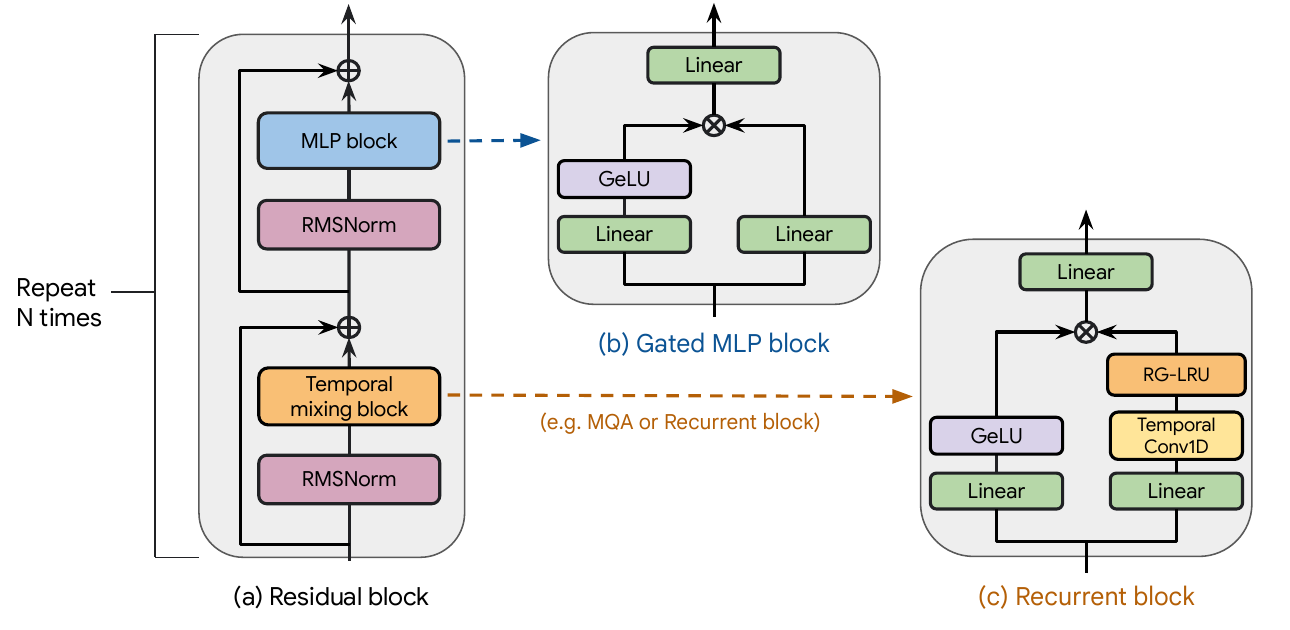}
\caption{
a) The main backbone of our mode architecture is the residual block, which is stacked $N$ times. 
b) The gated MLP \subblock{} that we use. %
c) The recurrent \subblock{} that we propose as an alternative to Multi Query Attention (MQA). 
It uses our proposed RG-LRU layer, defined in Section~\ref{sec:RGLRU}.}
\label{fig:model_sketch_1}
\end{figure}

The residual block contains two components, applied in order. 
The first component takes the hidden state $x$ and applies an RMSNorm \citep{zhang2019root}, followed by the temporal-mixing \subblock{}. 
We then merge the output with a skip connection from $x$ through addition.
Similarly, the second component applies RMSNorm, followed by the MLP \subblock{} and then merges its output with a skip connection from the input of the RMSNorm. This block is illustrated in Figure \ref{fig:model_sketch_1} (a). %

\subsection{MLP \subblock{}}

We use a gated MLP \subblock{} ~\citep{dauphin2017language} (illustrated in Figure~\ref{fig:model_sketch_1}(b)), which
creates two branches from its input of dimension $D$.
We apply a linear layer with output dimension $MD$ on each branch, where $M$ denotes the expansion factor. For simplicity, we use $M=3$ throughout this work. 
We apply a GeLU non-linearity \citep{hendrycks2016gaussian} on one of the branches before merging them 
by element-wise multiplication, similar to GeGeLU \citep{shazeer2020glu}.
However, in our MLP block, we apply a final linear layer with output dimension $D$ on the outputs of the GeGeLU layer. 

\subsection{Temporal-mixing \subblocks{}}

The temporal-mixing \subblock{} is the component of our model that aggregates hidden layer activations at different temporal locations in the sequence. 
We consider three temporal-mixing \subblocks{}: global MQA \citep{shazeer2019fast}, local MQA \citep{beltagy2020longformer} and our proposed \emph{Recurrent \subblock{}}.

\paragraph{Global multi-query attention}

Unless otherwise stated, we use MQA rather than MHA to improve the inference speeds of our Transformer baselines \citep{shazeer2019fast}. 
We use a fixed head dimension $D_{head} = 128$, and we fix the number of attention heads $H$ such that $HD_{head} = D$. This requires the model dimension $D$ to be a multiple of 128.
We do not use any absolute positional embeddings, but we use Rotary Position Embedding (RoPE) \citep{su2021roformer} as a relative positional embedding.

\paragraph{Local sliding window attention}
\label{sec:local_attention}

One of the key disadvantages of using global attention is that its computational complexity grows quadratically in the sequence length. 
To address this, several works have started to adopt \emph{local attention} \citep{beltagy2020longformer}, also known as sliding window attention. 
It allows each position to attend only to a fixed number of tokens in the past.
This not only reduces the computational FLOPs but also bounds the size of the KV cache to the size of window, making it no longer quadratic in the sequence length.
All other details are the same as the global MQA.

\paragraph{Recurrent \subblock{}}
\label{sec:recurrent_block}
Our recurrent \subblock{}  (Figure~\ref{fig:model_sketch_1}(c))  is similar to the GSS block \citep{mehta2022long} and the block used by Mamba \citep{gu2023mamba}.
We take the input of dimension $D$ and apply two linear layers with output dimension $D_{RNN}$ in parallel, creating two branches.
On the first branch, we apply a small separable Conv1D layer, inspired by the Shift-SSM in H3~\citep{dao2022hungry}, with a temporal filter dimension of 4.
Note that this Conv1D layer is very small, with just $4D_{RNN}$ parameters. 
We follow the Conv1D layer with our proposed RG-LRU layer (defined below.) %
On the second branch we apply a GeLU nonlinearity and then merge the branches by element-wise multiplication. 
We then apply a final linear layer with output dimension $D$.

\subsection{Real-Gated Linear Recurrent Unit (RG-LRU)}
\label{sec:RGLRU}
Our proposed RG-LRU layer has a simple recurrence inspired by the Linear Recurrent Unit (LRU) \citep{orvieto2023resurrecting}, but incorporates a gating mechanism motivated by the literature on non-linear RNNs, in particular LSTMs~\citep{hochreiter1997long} and GRUs~\citep{chung2014empirical}.
The equations describing the layer are as follows:

\begin{eqnarray}
 r_t &=& \sigma\left(W_a x_t + b_a\right), \quad \textcolor{gray}{\text{ \emph{recurrence gate}}} \label{eq:recurrent_gate} \\
 i_t & = & \sigma(W_x x_t + b_x), \quad \textcolor{gray}{\text{\emph{ input gate}}} \label{eq:input_gate}\\
a_t &=& a^{c r_t}, \label{eq:a_rglru} \\
h_{t} &=& a_t \odot h_{t-1} + \sqrt{1-a_t^2} \odot (i_t \odot x_t).
\label{eq:RG_LRU}
\end{eqnarray}
The output of the layer is $y_t = h_t$, and the non-linearity $\sigma$ in the equations is the $\sigmoid$ function. 
The recurrent weight $a$ in Equation \eqref{eq:RG_LRU} is diagonal. Hence all operations are element-wise.
We parameterize $a$ in Equation \eqref{eq:a_rglru} as $a = \sigma(\Lambda)$, where $\Lambda$ is a learnable parameter.
This guarantees that $0 \leq a \leq 1$, ensuring that the recurrence is stable.
The variable $c$ is a scalar-valued constant set to $8$.
For numerical stability, in practice we compute $a^{cr_t}$ in log-space (see Appendix~\ref{appendix:recurrence_gate}).
The layer has gates on both the input $x$ and the recurrent weight $a$. However, neither gate depends on the recurrent state $h_{t-1}$, which ensures that the computation can be executed efficiently on device. 
We initialize both $W_a$ and $W_x$ using LeCun init \citep{lecun2002efficient}. 
We initialize $\Lambda$ such that $a^c$ is uniformly distributed between $0.9$ and $0.999$ at the start of training, similar to \cite{orvieto2023resurrecting}. 
Unlike many recent works in the SSM literature, the RG-LRU does not use initialization inspired by the theory of orthogonal polynomials \citep{gu2020hippo}, and it also is not defined as the discretization of an underlying continuous system \citep{gu2021efficiently}. 
Unlike the original LRU layer, we do not use complex algebra in the recurrence. While using complex recurrences would lead to a more expressive layer~\citep{orvieto2023universality} we found that complex recurrences were not beneficial for language modelling in practice, as also observed by \citet{gu2023mamba}.\footnote{We suggest ablating the use of complex numbers for other modalities and provide more information about the complex-valued version of the RG-LRU layer in Appendix~\ref{appendix:complex_lru}.} %

\paragraph{Gate behaviour}

The \emph{input gate} $i_t$ is similar to the one in LSTM, which can filter (or scale down) the input $x_t$.
However, to our knowledge, our recurrence gate $r_t$ is different from other gating mechanisms in the literature.
For example, the \emph{selection mechanism} proposed in Mamba  \citep{gu2023mamba} is comparable to the \emph{update gate} of GRUs which interpolates between the previous state and and the current observation $x_t$.
Its effect on the hidden state allows it to \emph{reset its state and forget any information it holds from the past}, similar to the forget gate in the LSTM. 
In contrast, our recurrence gate can approximately interpolate between the standard LRU update from \citet{orvieto2023universality} and the previous hidden state, which allows it to effectively \emph{discard the input and preserve all information from the previous history} (see Appendix~\ref{appendix:recurrence_gate} for further details). We believe the key role of this gate is to enable the model to achieve super-exponential memory by reducing the influence of uninformative inputs.

\section{Recurrent Models Scale as Efficiently as Transformers}
\label{sec:scaling_new}
Scaling studies provide important insights into how to tune the hyperparameters of a model and its behaviour at scale. 
Here, we define the models evaluated in our studies, and provide scaling curves up to and beyond 7B parameters. Finally, we assess the performance of our models on downstream tasks.
We consider 3 model families in this work; (1) a MQA-Transformer baseline, (2) Hawk; our pure RNN model, and (3) Griffin; our hybrid model which mixes recurrent blocks with local attention.  We define the key model hyper-parameters for models across a range of scales in Appendix \ref{appendix:model_scale_hparams}. %

\paragraph{MQA Transformer baseline}
\label{sec:MQA-Transformer-baseline}

Our Transformer baseline uses the residual pattern and the gated MLP block described in Section \ref{sec:model}, in combination with MQA \citep{shazeer2019fast} and RoPE \citep{su2021roformer}.

\paragraph{Hawk}
\label{sec:hawk}
The Hawk architecture uses the same residual pattern and MLP block as our Transformer baseline, but we use the recurrent block introduced in Section~\ref{sec:recurrent_block} with a RG-LRU layer (see Section~\ref{sec:RGLRU}) as our temporal mixing block, instead of MQA. 
We expand the width of the recurrent block by a factor of approximately $\frac{4}{3}$ (i.e. $D_{RNN} \approx 4D/3$) in order to roughly match the parameter count of a MHA block when both use the same model dimension $D$.\footnote{Note that we match parameters with MHA attention block, though our Transformer baseline and Griffin ended up relying on MQA attention in order to improve inference efficiency. This means that our recurrent blocks have slightly more parameters than the corresponding MQA blocks.} See Appendix \ref{appendix:model_scale_hparams} for precise hyper-parameters.

\paragraph{Griffin}
\label{sec:griffin}

The key advantage of recurrent blocks over global attention is that they use a fixed state size to summarize the sequence, whereas the size of MQA's KV cache grows proportional to sequence length. 
Since local attention (Section~\ref{sec:local_attention}) has the same property, mixing recurrent blocks with local attention preserves this benefit. 
We have found this combination extremely effective, since local attention accurately models the recent past, while the recurrent layers can transmit information across long sequences.

Griffin uses the same residual pattern and MLP block as our Transformer baseline. However unlike both our MQA Transformer baseline and the Hawk model, Griffin uses a mixture of recurrent blocks and MQA blocks.
Specifically, we employ a layered structure by alternating two residual blocks with a recurrent block followed by one residual block which uses the local (MQA) attention block described in Section~\ref{sec:local_attention}. 
Unless otherwise stated, the local attention window size is fixed to 1024 tokens.

\subsection{Scaling curves}

We present our main scaling results in Figure~\ref{fig:intro_a}. 
All three model families are trained at a range of model scales from 100M to 7B parameters, with an additional Griffin model with 14 billion parameters.
We increase the number of training tokens to be roughly proportional to the number of parameters of the model, as prescribed by the Chinchilla scaling laws \citep{hoffmann2022training}. 
Models are trained on the MassiveText dataset \citep{hoffmann2022training}, previously used to train Gopher \citep{rae2021scaling} and Chinchilla \citep{hoffmann2022training}, although we use a slightly different data subset distribution.
A sequence length of 2048 tokens was used (see Section~\ref{sec:long_seq} for results with longer sequences.) 
All experiments use the AdamW optimizer \citep{loshchilov2017decoupled}. We tune the learning rate, weight decay and $\beta_2$ parameters for small models, and use these runs to identify scaling rules for these hyper-parameters which predict their optimal values for the 7B and 14B models.

All three model families demonstrate a linear scaling relationship between the validation loss and training FLOPs (see Figure~\ref{fig:intro_a}; note both axes are in log scale), as previously observed for Transformers by \citet{brown2020language}. 
Notably, Griffin achieves lower validation loss than the Transformer baseline across all FLOPs budgets despite not using any global attention layers. Hawk on the other hand achieves slightly higher validation loss, but this gap appears to close as the training budget increases.

\subsection{Evaluation on downstream tasks}
\label{sec:downstream_evals}

In order to compare to other models in the literature, we train all our models for 300B tokens before evaluating on downstream tasks. 
The two external baselines that we compare to are Mamba-3B \citep{gu2023mamba}, the strongest small recurrent model reported in the literature to date, and Llama-2 \citep{touvron2023llama}, a widely used open Transformer model. Both external baselines have been trained on significantly more than 300B tokens -- Mamba has been trained on 600B tokens, twice more, and Llama-2 has been trained on 2T tokens, nearly seven times more. We note however that both Mamba and Llama-2 were trained on different datasets and with different hyper-parameter tuning strategies, which may partially explain our strong performance. We therefore also include our own MQA transformer baseline, trained on the same data and with the same hyper-parameter tuning budget as Hawk and Griffin.

We provide an evaluation on downstream tasks in Table \ref{tab:full_evals}. 
We find that both Hawk and Griffin achieve very strong performance. 
In line with other works, we report character normalized accuracy on MMLU, HellaSwag, PIQA, ARC-E and ARC-C, while we report absolute accuracy on WinoGrande with partial scoring. The performance of Hawk improves significantly as we increase the model size, and Hawk-3B achieves stronger performance on downstream tasks than Mamba-3B, despite being trained on half as many tokens. 
Griffin-3B significantly outperforms Mamba-3B, and Griffin-7B and Griffin-14B achieve performance competitive with Llama-2, despite being trained on nearly 7 times fewer tokens. Hawk is also competitive with our MQA Transformer baseline, while Griffin outperforms this baseline.

\begin{table}[t!]
\caption{Character normalized accuracy. Hawk is competitive with our Transformer baseline, and exceeds the reported performance of Mamba despite being trained on half as many tokens. Griffin outperforms our Transformer baseline, and matches the performance of Llama-2 despite being trained on roughly 7 times fewer tokens. We report unnormalized accuracy with partial scoring for WinoGrande.}
\renewcommand{\arraystretch}{1.15}  %
\resizebox{\textwidth}{!}{%
\begin{tabular}{lll|cccccc|c}
 \toprule
     \textbf{Model}
     & \textbf{Model} 
     & \textbf{Training} 
     & \multirow{2}{*}{\textbf{MMLU}} 
     & \multirow{2}{*}{\textbf{HellaSwag}} 
     & \multirow{2}{*}{\textbf{PIQA}} 
     & \multirow{2}{*}{\textbf{WinoGrande}} 
     & \multirow{2}{*}{\textbf{ARC-E}} 
     & \multirow{2}{*}{\textbf{ARC-C}} 
     & \multicolumn{1}{l}{ \multirow{2}{*}{\textbf{Average}}} 
 \\
    \textbf{Type} 
    & \textbf{Size} 
    & \textbf{Tokens} 
    &  &  &  &  &  &  & \\ \midrule
\textbf{Mamba} & 3B & 600B & 26.2 & 71.0 & 78.1 & 65.9 & 68.2 & 41.7 & 58.5 \\ %

\rowcolor{tablerowcolor}  & 7B & 2T & 45.3 & 77.2 & 78.8 & 69.2 & 75.2 & 45.9 & 65.3 \\
\rowcolor{tablerowcolor} \multirow{-2}{*}{\textbf{Llama-2}} & 13B & 2T & \textbf{54.8} & 80.7 & 80.5 & 72.8 & 77.3 & 49.4 & 69.3 \\ \midrule

\textbf{MQA} & 1B & 300B & 28.9 & 64.8 & 75.0 & 62.0 & 60.2 & 35.4 & 54.4 \\
 \textbf{Transformer} & 3B & 300B & 31.7 & 71.0 & 77.6 & 66.1 & 68.1 & 39.2 & 59.0 \\
\textbf{(Baseline)} & 6B & 300B & 38.9 & 77.0 & 79.5 & 70.4 & 74.1 & 45.2 & 64.2 \\ %

\rowcolor{tablerowcolor}  & 1B & 300B & 29.7 & 63.3 & 76.1 & 57.2 & 60.6 & 34.6 & 53.6 \\
\rowcolor{tablerowcolor} & 3B & 300B & 31.3 & 71.7 & 78.8 & 66.5 & 68.4 & 40.2 & 59.5 \\
\rowcolor{tablerowcolor} \multirow{-3}{*}{\color{tableemphcolor} \textbf{Hawk}} & 7B & 300B & 35.0 & 77.6 & 80.0 & 69.9 & 74.4 & 45.9 & 63.8 \\ %

\multirow{4}{*}{\color{tableemphcolor} \textbf{Griffin}} & 1B & 300B & 29.5 & 67.2 & 77.4 & 65.2 & 67.0 & 36.9 & 57.2 \\
 & 3B & 300B & 32.6 & 73.5 & 78.1 & 67.2 & 71.5 & 41.4 & 60.7 \\
 & 7B & 300B & 39.3 & 78.6 & 81.0 & 72.6 & 75.4 & 47.9 & 65.8 \\
 & 14B & 300B & 49.5 & \textbf{81.4} & \textbf{81.8} & \textbf{74.1} & \textbf{79.1} & \textbf{50.8} & \textbf{69.5}
\end{tabular}
} 
\label{tab:full_evals}
\end{table}

\section{Training Recurrent Models Efficiently on Device}
\label{sec:training_efficiency}

We encountered two main engineering challenges when developing and scaling our models.
First, how to efficiently shard our models across multiple devices. 
Second, how to efficiently implement linear recurrences to maximize training efficiency on TPUs. 
We address both of these challenges in this section, before providing an empirical comparison of the training speed of Griffin and our MQA baseline.

\subsection{Model parallelism for large scale training}

As our model increases in size, we cannot fit the model on a single device during training, even with a batch size of 1 per-device. 
We therefore use model parallelism to shard our large models across devices during training. 
Since communication costs across different training devices are expensive, efficiently sharding the model is critical for fast training at scale. 

\paragraph{MLP and MQA block} 
For our gated-MLP block we use Megatron-style sharding \citep{shoeybi2019megatron}, which requires  a single all-reduce operation in both the forward and the backward pass. 
Similarly, we apply the same strategy to the linear layers in the attention block, and additionally shard the attention mechanism over its heads  \citep{narayanan2021efficient}.

\paragraph{Recurrent Block}

The recurrent block contains two linear layers per branch. This allows us to apply Megatron sharding to these layers in an equivalent fashion. The Conv1D layer operates independently across channels, enabling us to split its parameters across devices without incurring any communication overhead. 
To avoid additional cross-device communication, we use block-diagonal weights for the gates in the RG-LRU (see equations~\ref{eq:recurrent_gate} and~\ref{eq:input_gate}), instead of dense matrices. For all experiments in this paper, we use 16 blocks for both the recurrence gate and the input gate (such that $W_x$ and $W_a$ each have $D_{RNN}^2/16$ parameters).
The diagonal structure of the recurrence offers the same advantage as the Conv1D, allowing parameter sharding and computation without any communication. %
With this strategy, the recurrent block's communication requirements are equivalent to those of the MLP block.

\paragraph{Other considerations}

Optimizer states can consume significant memory, exceeding the size of the model parameters themselves. 
To address this, we employ ZeRO parallelism \citep{rajbhandari2020zero}, distributing both optimizer states and model parameters across the batch shards. 
We also use bfloat16 representation for model parameters and activations, minimizing any data transfer overhead.

\subsection{Efficient linear recurrences on device}
\label{sec:training_rnn}

Current deep learning accelerators are optimized for classical architectures which are composed largely of matrix multiplications and convolutions. 
These operations have a high FLOPs-to-byte ratio, motivating the development of specialized hardware units like Nvidia GPUs' TensorCores  \citep{markidis2018nvidia} and Google TPUs' MXUs \citep{norrie2021design, jouppi2021ten, jouppi2023tpu}. 
Classical RNNs also benefit from this due to their dense recurrence matrices.
In contrast, our proposed RG-LRU layer, like other diagonal RNN models, has a low FLOPs-to-byte ratio.
This fundamental difference poses a computational challenge, as existing accelerators lack optimization for such workloads. 
Since we run all our experiments on TPU-v3, we focus on developing an efficient implementation tailored to this device\footnote{The conclusions drawn here do not necessarily apply to other accelerators.}.

\paragraph{Challenges for linear diagonal RNNs}

One of the main challenges of utilizing a device like the TPU-v3 for the RG-LRU is that the update equation of the hidden state in eq.~(\ref{eq:RG_LRU}) is a pure elementwise operation.
For each element update it requires to load 6 bytes (assuming bfloat16 we need 2 bytes for each of the variables $h_{t-1}, a_t, x_t$) and write 2 bytes (the hidden state $h_t$) while the computation only executes 6 FLOPs (number of arithmetic operations in eq.~\ref{eq:RG_LRU}) per element. 
This translates to a low FLOPs-to-byte ratio of $0.75$ -- significantly below the device's capacity for elementwise operations of $4.2$ (see Appendix~\ref{appendix:tpu_specs}).
Execution time is therefore dominated by memory transfers between HBM and VMEM, making the computation memory bound.

\paragraph{A custom linear scan}

To address this we have written a custom \href{http://go/jax-pallas}{Pallas} kernel for the computation of eq.~(\ref{eq:RG_LRU}) using a \emph{linear scan}. 
This allows us to minimize memory transfers, by keeping the hidden state in VMEM all the time, and also to perform memory transfers in larger chunks rather than one at a time. 
In practice, this translates to almost 3x speed up over the native Jax implementation of the linear scan. 
Additionally, we observe 10-20\% lower training times per step of the full Hawk model, relative to the same model using the native Jax implementation (see Appendix~\ref{sec:appendix_scan} for more details.)

\paragraph{Why we do not use convolutions or associative scans?}

The initial appeal of linear recurrence models stemmed from their high parallelizability, enabled by the associativity of their computations. 
This permitted efficient execution on device via convolutions \citep{gu2021combining} or prefix-sum algorithms (the associative scan) \citep{smith2022simplified}. 
However, the RG-LRU's gating mechanism on $a_t$ is not compatible with the convolutional view.
Although we can still use the associative scan in principle, the associative scan reduces the number of FLOPs required but does not reduce memory overheads, which is our primary bottleneck in practice. 
Empirically we observe that on a TPU-v3 the associative scan is significantly slower that the native Jax linear scan (see Appendix~\ref{sec:appendix_scan} for more details.)
We speculate that the random access nature of the tree recombination of the parallel prefix-sum algorithm makes is poorly suited for the TPU architecture, leading to even slower memory transfers -- the main bottleneck of this operation.

\subsection{Training speed on longer sequences}

\begin{figure}[t]
\centering
\subfigure[400m]
{\includegraphics[width=0.3\linewidth]{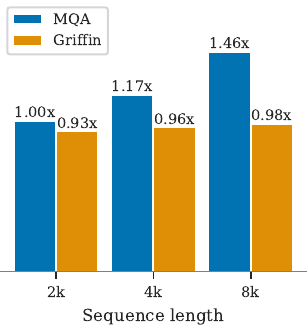}
\label{fig:longer_seq_a}}
\subfigure[1B]
{\includegraphics[width=0.3\linewidth]{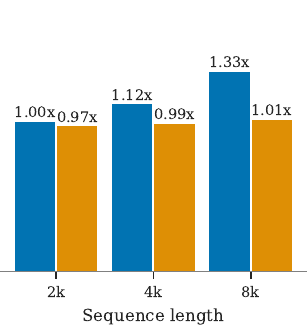}
\label{fig:longer_seq_b}}
\subfigure[7B]
{\includegraphics[width=0.3\linewidth]{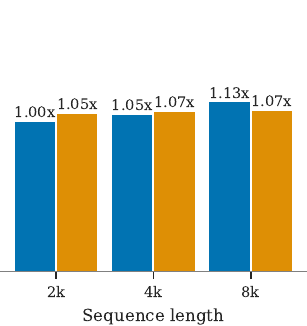}
\label{fig:longer_seq_c}}
\caption{
Training durations per step computed relative to our MQA baseline at 2K sequence length as we vary the model size and sequence length for Griffin and MQA. 
Let us note that as we increase the sequence length we lower the batch size proportionally, such that the total number of tokens per batch stays fixed.
}
\label{fig:longer_seq}
\end{figure}

We compare the training speeds across different model sizes and sequence lengths to investigate the computational advantages of our models during training. 
For each model size, we keep the total number of tokens per batch fixed, meaning that as we increase the sequence length, we proportionally decrease the number of sequences. 
In Figure~\ref{fig:longer_seq}, we plot the relative runtimes of our Griffin model compared to that of the MQA baseline at 2048 sequence length. 
At the lowest sequence length, the two models have similar training time, but as we increase the sequence length the Transformer becomes slower, while Griffin's runtime remains the same. 
The drop in speed for the baseline is more pronounced at smaller model sizes and decreases at larger model sizes. 
This can be explained by the fact that all models contain a large number of linear layers.
Their computation scales $O(T D^2)$, while the RG-LRU  is $O(T D)$ vs $O(T^2 D)$ of global attention. 
This means that as we increase the model width $D$ compared to the sequence length $T$, the linear layers become the primary computational bottleneck, minimizing the efficiency gains from the RNN \subblock{}.
Therefore, replacing Transformers with Hawk or Griffin offers the most significant wall-clock time improvement when sequence length is sufficiently large relative to model width to ensure the attention computation constitutes a major portion of the total computation time. We also note that in practice, our MQA baseline has slightly fewer parameters than Griffin at the same model scale (and performs fewer FLOPs). This explains why Griffin trains slightly slower than our MQA baseline at 7B for short sequences.

\section{Inference Speed}
\label{sec:inference}

Inference in LLMs is composed of two stages. In the ``prefill'' stage, we receive and process the prompt. This step is effectively performing a forward pass of the model. 
Since the prompt can be processed in parallel across the sequence, most model operations are compute bound during this stage. 
We therefore expect the relative speeds of Transformers and recurrent models during the prefill stage to be similar to the relative speeds of the same models during training, which we discussed in Section \ref{sec:training_efficiency}. 

Prefill is followed by a ``decode'' stage, in which we sample tokens auto-regressively from the model. As we show below, recurrent models have lower latency and higher throughput during the decoding stage, especially for longer sequence lengths where the key-value (KV) cache used in attention can get large.

There are two main metrics to consider when evaluating inference speed. 
The first is latency, which measures the time taken to generate a specified number of tokens at a certain batch size. 
The second is throughput, which measures the largest number of tokens per second that can be generated on a single device when sampling a specified number of tokens. Since throughput is given by tokens sampled times batch size divided by latency, one can improve throughput either by reducing the latency or by reducing memory usage to enable the use of larger batch sizes on device. Latency can be useful to consider for real-time applications that require a quick response time. Throughput is also useful to consider as it can tell us the maximum number of tokens we could sample from a particular model in a given time. This property is useful when considering other language applications such as Reinforcement Learning from Human Feedback (RLHF) or scoring language model outputs such as done in AlphaCode \citep{li2022competition} where being able to output a large number of tokens in a given time is an appealing feature.

\subsection{A simple model of the decode step}

All components of language models are memory bound during decoding as long as batch size isn't too big (i.e. $B \lesssim 128$- see Appendix \ref{appendix:memory_bound} for details) and we will assume this for the remainder of this section.
The largest memory overheads of Transformers typically come from the parameters themselves and the KV cache. 
Therefore we can approximate the time required to generate a single token for each sequence in the batch $B$ during decoding as the time needed to load these two quantities from memory:
\begin{equation}
    \textit{Time to sample next token} \approx \frac{\textit{param size} + \textit{batch size} \times \textit{cache size}}{\textit{memory bandwidth}}.
\end{equation}
Here, \textit{cache size} refers to either the size of the KV cache at batch size 1 (for Transformers), or to the size of the recurrent state at batch size 1 (for RNNs).

\paragraph{Cache sizes}

The difference in cache size relative to model parameters has important implications for sampling efficiency.
In recurrent and local attention blocks, parameter loading is the primary bottleneck, (because the cache size is substantially smaller).
In contrast, global attention's KV cache scales with the sequence length $T$ and can be comparable to, or even exceed, the size of the model parameters.
 This introduces considerable overhead when the sequence length $T$ is large enough (as shown in \ref{appendix:cache_sizes}).
Consequently, an equally sized recurrent model can exhibit substantially lower latency than a Transformer when $T$ is large. Note however that as the model size grows the sequence length at which we see latency benefits (where the KV cache size is comparable to parameter size) also increases.
It is important to note that, as well as improving latency, having a small recurrent state can also increase the largest batch size that fits in memory on a single device, leading to higher throughput.

\begin{figure}[t]
\centering
\subfigure[Latency empty prefill]{
\includegraphics{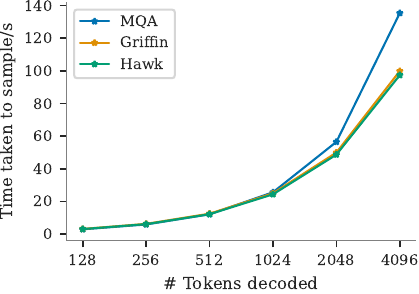}\label{fig:latency_empty_prefill}
}
\subfigure[Latency 4k prefill]{
\includegraphics{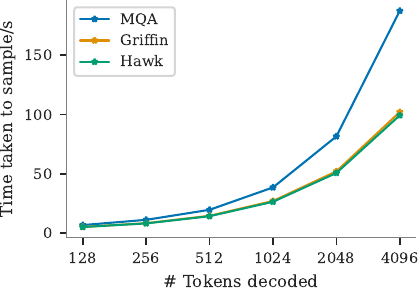}\label{fig:latency_4k_prefill}
}
\caption{Latency of different 1B parameter models for a range of sequence lengths for (a) sampling from an empty prefill and (b) sampling from a prefill of 4k tokens.}
\label{fig:latency}
\end{figure}

\subsection{Results}
Here, we look at inference results for models of size 1B parameters. For our baseline, we compare against a MQA Transformer, which is significantly faster during inference than the standard MHA Transformer often used in the literature. The models that we compare are: i) \textit{MQA Transformer}, ii) Hawk, and iii) Griffin. For comparing different models we report both latency and throughput.

\paragraph{Latency}
We compare the latency for models with a batch size of 16 with an empty prefill as well as a prefill of 4096 tokens as seen in Figure~\ref{fig:latency}. 
Hawk and Griffin achieve faster sampling latency than MQA Transformers for long sequences. This is particularly noticeable as the sequence length and the prefill length (which affect the size of the KV cache) are increased. Griffin achieves similar latency to Hawk, demonstrating the excellent compatibility of linear recurrences and local attention.

\paragraph{Throughput}
We compare the maximum throughput (tokens/s) for the same models when sampling 512, 1024, 2048 and 4196 tokens following an empty prompt in Figure~\ref{fig:intro_b}. 
We see that both Griffin and Hawk achieve significantly higher throughput than the MQA Transformer baseline. 
This is partially due to recurrent models having lower latency but also primarily occurs because Griffin and Hawk can fit larger batch sizes than the MQA Transformer on a single device, since their cache size is smaller. Hawk achieves higher throughputs than Griffin, since the size of the local attention cache eventually becomes comparable to the size of the parameters when the batch size is large.

\section{Long Context Modeling}
\label{sec:long_seq}

In this section, we explore the effectiveness of Hawk and Griffin to use longer contexts to improve their next token prediction, and investigate their extrapolation capabilities during inference. 
Additionally, we explore our models' performance on tasks that require copying and retrieval capabilities, both for models that are trained on such tasks, as well as when testing for these capabilities with our pre-trained language models.

\subsection{Improving next token prediction with longer contexts}

We investigate the ability of Hawk and Griffin to improve their predictions with longer contexts. In particular, we evaluate our trained models by measuring the loss on a held-out books dataset across a range of sequence lengths. Using these long documents allows us to evaluate the ability of the models to extrapolate, i.e. the ability to accurately predict the next token given contexts that are longer than those seen during training.

\begin{figure}[t]
\centering
\includegraphics{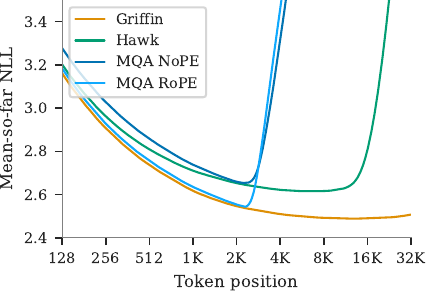}
\includegraphics{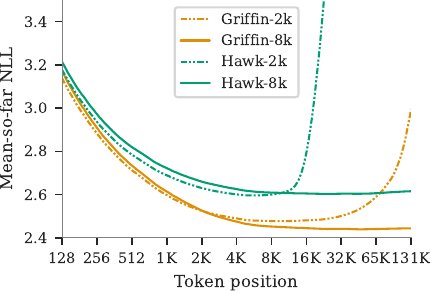}
\caption{
    Performance of various 1B parameter models on a held-out evaluation set of books.
    On the left, the models have been trained with sequence length 2048, and on the right with sequence lengths of respectively 2048 (2k) and 8192 (8k).
    Hawk and Griffin are able to extrapolate to significantly longer sequences than the Transformer baselines, and further improve performance when trained on longer sequences.
}
\label{fig:longseq}
\end{figure}

In Transformers, this ability to extrapolate is largely determined by the positional encoding used for the attention layers \citep{kazemnejad2024impact}.
For recurrent models, it is instead dictated by the capacity of the model to keep refining the representation stored in the recurrence state as the context becomes longer. 
From the left plot of Figure~\ref{fig:longseq}, we observe that, up to some maximal length, both Hawk and Griffin improve next token prediction given longer contexts, and they are overall able to extrapolate to significantly longer sequences (at least 4x longer) than they were trained on. 
In particular, Griffin extrapolates remarkably well even when using RoPE \citep{su2021roformer} for the local attention layers.

The results so far evaluate models that have been trained on sequences of 2048 tokens. 
In order to assess whether our models can also effectively learn from longer contexts, we train 1B parameter models on sequences of 8192 (8k) tokens on MassiveText, and compare them to models trained on the same dataset but on sequences of length 2048 (2k) tokens. 
We keep the total number of training tokens the same across the models by reducing the batch size by a factor of 4 for the models trained on the sequence length of 8192 (while keeping the number of training steps fixed).
As illustrated in the right plot of Figure~\ref{fig:longseq}, we find that Hawk-8k and Griffin-8k do achieve lower evaluation loss for sequences of length 8192 or larger, compared to Hawk-2k and Griffin-2k. 
This indicates that Hawk and Griffin are able to learn to use longer contexts during training. 
Interestingly, when evaluating at short sequence lengths, we find that Hawk-2k and Griffin-2k perform slightly better than Hawk-8k and Griffin-8k. 
This suggests that the training sequence length should be carefully chosen according to the intended downstream use of the model.

\subsection{Copy and retrieval capabilities}

Recent work \citep{jelassi2024repeat} has shown that Transformers can be significantly more efficient than state space models (SSMs), a popular new family of RNNs, at learning synthetic tasks such as copying the context or retrieving relevant tokens from the context. Additionally, \citet{jelassi2024repeat} showed that pre-trained Transformers such as Pythia \citep{biderman2023pythia} are much better at copying and retrieval tasks at evaluation time compared to pre-trained SSM models such as Mamba \citep{gu2023mamba}. In this section, we investigate the efficiency of Griffin and Hawk in learning how to copy and retrieve tokens from the context. Additionally, we evaluate pre-trained Hawk and Griffin models on a phone number lookup task designed to test both copying and retrieval capabilities.

\begin{figure}[t]
\centering
\subfigure[Selective Copying Task]{\includegraphics{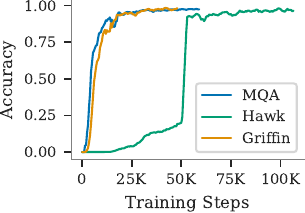}}
\hspace{.25cm}
\subfigure[Induction Heads Task]{\includegraphics{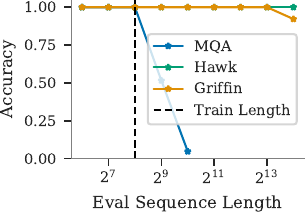}}
\hspace{.25cm}
\subfigure[Phonebook Lookup Task]{\includegraphics{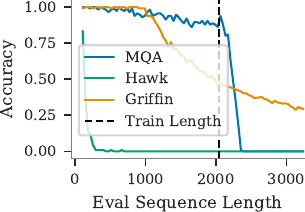}}
\caption{Exploring the copying and retrieval capabilities of Hawk and Griffin on three synthetic tasks. Figures (a) and (b) show the performance of 5 layer deep models on a held out eval set when explicitly trained on these tasks. Figure (c) shows the performance on a phone number lookup task when evaluating our pre-trained 7B Hawk and Griffin models against our 6B MQA Transformer baseline.}
\label{fig:synthetic_tasks}
\end{figure}

\paragraph{Training on synthetic tasks} 

To investigate the efficiency of learning how to copy and retrieve relevant tokens from the context, we train on two synthetic tasks: Selective Copying and Induction Heads. To be able to compare Transformers with Hawk and Griffin, we consider 5-block deep networks with model dimension 64, totalling roughly 250K parameters, where Griffin uses a single local attention in the middle of the network, in the third block.
\begin{itemize}
    \item \textbf{Selective copying task}: In this task, the model needs to learn to copy data tokens from a sequence while ignoring noise tokens from the context. See Appendix \ref{app:synthetic_tasks} for more details on the setup for this task. This task is inspired by \citet{gu2023mamba}, where the authors showed that Mamba was able to solve this task better than previously proposed SSMs. We use a vocabulary size of 16, and train on sequences of length 1024, containing 16 data tokens (randomly sampled from the vocabulary and at random locations), with the rest of the tokens set to the noise token. Griffin uses a local attention window size of 512. \\
    \item \textbf{Induction heads}: In this task, the model needs to learn to recall the token immediately following a special token. This requires the model to learn the special token, and retrieve the token immediately following it in the context. If the model is able to learn the task, it should be able to extrapolate to significantly longer sequences than it was trained for. We use a vocabulary size of 16 and train on sequences of length 256 where the tokens are sampled randomly, and we randomly sample the location of the special token in the sequence. Griffin uses a local attention window of size 128.
\end{itemize}

We show our results in Figure \ref{fig:synthetic_tasks}. 
On the Selective Copying task, we find that all 3 models are able to solve the task perfectly. 
When comparing speed of learning on this task, we find Hawk to be significantly slower than Transformers, similar to the observation made by \citet{jelassi2024repeat}, where the authors showed that Mamba was significantly slower to learn on similar tasks. 
Interestingly though, Griffin shows almost no slowdown, effectively matching the speed of learning of Transformers, despite using only a single local attention layer.

On the Induction Heads task, while all 3 models can solve the task perfectly up to the training sequence length, our Transformer baseline is not able to extrapolate to longer sequences during evaluation. 
While our MQA baseline uses RoPE, \citet{gu2023mamba} had similar observation for Transformers with a range of positional encodings. 
We find that Hawk is able to perfectly extrapolate on this task to evaluation sequences several orders of magnitude longer than the training sequence length. 
Notably, Griffin, with its local attention, also demonstrated exceptional ability to extrapolate on this task.

\paragraph{Evaluating pre-trained models} 

We now evaluate whether copying and retrieval capabilities naturally emerge in our pre-trained models. We consider our 7B Hawk and Griffin models and our 6B MQA Transformer baseline, all trained on 300B tokens on the MassiveText dataset. 
We consider the same phonebook lookup task introduced in \citet{jelassi2024repeat}, where we provide to the model a synthetic phonebook containing names and numbers, and the model is asked to retrieve the correct phone number given a name. 
The prompt to the model is a phonebook consisting of randomly sampled list of names and numbers of a certain length, followed by two randomly sampled examples of the task, followed by a randomly sampled name from the phonebook for which the model needs to retrieve the correct phone number.

From Figure \ref{fig:synthetic_tasks}(c), we see that while Hawk can do reasonably well on the task for very short phonebook lengths, it fails to memorize and retrieve the correct phone number when the phonebook length grows, similar to the observation made by \citet{jelassi2024repeat} on the Mamba model's performance on this task. This is not particularly surprising since Hawk uses a small fixed-size state. Our Transformer baseline can almost perfectly solve this task up to the training sequence length, but fails to retrieve the correct phone number for context lengths longer than the training sequence length. Interestingly, Griffin can perfectly solve this task up to a context length that matches its local attention window size of 1024, in spite of using only a single local attention layer. Once the context length is long enough such that the local attention window does not cover the whole phonebook, performance starts to degrade. Griffin is also able to extrapolate better to longer sequence lengths compared to Transformers. 
While the performance of Griffin is promising for the ability of models with fixed-size state to solve copying and retrieval tasks, our results suggest more work is needed to improve these  capabilities for such models.

\section{Related Works}

The Transformer architecture has become a more scalable alternative to RNNs. 
Transformers achieve superior scalability through fully parallelized training, contrasting with the inherent limitations of RNNs. 
Due to their sequential processing structure, classical RNNs suffer from slow training speeds during both forward and backward propagation \citep{werbos1990backpropagation}. 
To mitigate this issue, researchers have explored alternative RNN-based methods. 
Notable examples include Quasi-RNNs \citep{bradbury2016quasi}, which combine convolutions and linear RNNs for greater parallelization, and the use of input-based gating mechanisms to parallelize linear RNN training \citep{martin2017parallelizing}.

State-space Models (SSMs) have recently emerged as a powerful tool for modeling long input sequences.
They demonstrated strong performance on tasks from the long-range arena benchmark \citep{tay2020long}, and audio generation \citep{goel2022s}.
SSMs successfully integrate concepts from classical state-space models \citep{kalman1960new} with those of RNNs. 
Their reliance on linear recurrences allows for efficient hidden state computation, either through parallel scan operations or convolutions, resulting in training speeds comparable to Transformer models. 
The S4 \citep{gu2021efficiently} model proposed a sophisticated parameterization called \textbf{normal plus low-rank} to diagonalize the recurrence computation.
The S4D parametrized the SSM directly with a diagonal state matrix and showed that it performed just as well while being much simpler \citep{gu2022parameterization}. S5 also diagonalized the recurrence, and showed that the recurrence can be computed using the associative scan \citep{smith2022simplified}.
The H3 model \citep{dao2022hungry} generalizes the recurrent interpretation of linear attention \citep{katharopoulos2020Transformers}.
Hyena \citep{poli2023hyena} uses a similar architecture, but replaces the S4D layer with a global convolution kernel parametrized by an MLP. 
RetNet \citep{sun2023retentive} uses a simpler SSM design with a gating mechanism which allows them to parallelize the computation using a variant of multi-head attention. 
\cite{orvieto2023resurrecting}  systematically analyzed and ablated multiple modifications to standard RNNs.
Their finding showed that through better parameterization and initialization simplified linear RNNs (the LRU), perform just as well as other SSMs variants on various long-range tasks. 
RWKV \citep{peng2023rwkv} is a recent RNN, shown to be competitive on language modeling tasks, based on another linear attention approximation inspired by the attention-free Transformer \citep{attention-free-Transformer}.
Concurrent to our work \cite{gu2023mamba} developed an SSM architecture called Mamba with an input dependant selection mechanism and showed that it achieves performance comparable to Transformers with efficient inference. Several extensions of Mamba have been proposed \citep{wang2024mambabyte,zhu2024vision} for different applications. An input-dependent gating similar to Mamba was also proposed by Gateloop \citep{katsch2023gateloop}.

Linear attention \citep{katharopoulos2020Transformers} offers a computationally efficient approximation of the self-attention mechanism by linearizing the attention, which 
can be computed recurrently as a linear RNN.
While this approach significantly reduces computational cost compared to full attention, it often comes with a trade-off in model performance. 
Flash Attention \citep{dao2022flashattention} improves the training speed of attention on GPUs by making efficient use of the memory hierarchy. 
Another approach to reducing the computational cost of global attention, which is becoming increasingly more popular, is using sparse-local attention \citep{child2019generating} or sliding window attention \citep{jiang2023mistral}.

\section{Conclusion}

This work introduces Hawk; a recurrent model incorporating a novel gated linear recurrent layer, the RG-LRU. We also introduce Griffin; a hybrid model which mixes the RG-LRU layer with local attention.
These models demonstrate exceptional language modeling performance across varying scales, with held-out loss exhibiting power-law scaling as compute resources increase. Hawk exceeds the reported performance of Mamba on downstream tasks when trained on half as many tokens, while Griffin slightly exceeds the performance of Llama-2 when trained on over 6 times fewer tokens.
Furthermore, we empirically validate the inference-time advantages of Hawk and Griffin and observe reduced latency and significantly increased throughput compared to our Transformer baselines.  
Lastly, Hawk and Griffin exhibit the ability to extrapolate on longer sequences than they have been trained on and are capable of efficiently learning to copy and retrieve data over long horizons.
These findings strongly suggest that our proposed models offer a powerful and efficient 
alternative to Transformers with global attention.

\section*{Acknowledgements}
We thank Adam Paszke, Sharad Vikram, Trevor Gale, Sebastian Borgeaud, George Scrivener, Raia Hadsell, Oriol Vinyals, Toby Boyd, Zhifeng Chen, Chris Dyer, Kelvin Xu, Andriy Mnih for their guidance and advice. 
We make use of the DeepMind Jax ecosystem \citep{jax2018github} and especially thank Andy Brock for building the internal framework we used for training and evaluating our models.

\bibliography{main}

\newpage
\appendix
\section{RG-LRU Recurrence Gate}
\label{appendix:recurrence_gate}
\begin{figure}[htbp!]
\centering
\includegraphics[width=0.95\linewidth]{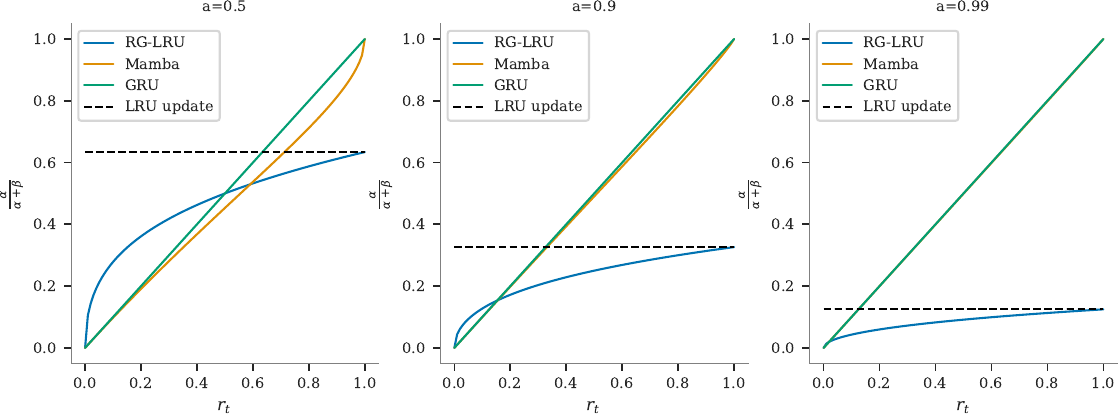}
\caption{The behaviour of different gating mechanisms applied on the recurrent weight $a$ (note that in the Mamba's notations this is $-A$).}
\label{fig:different gates}
\end{figure}

In Figure \ref{fig:different gates}, we demonstrate the behavior of different gating mechanisms applied on the recurrent weight $a$.

\paragraph{Implementation}

We implement our recurrence gate, as defined in Section~\ref{sec:RGLRU}, in a slightly different, but mathematically equivalent form, for numerical stability. 
In particular, we compute the logarithm of $a_t$ and then we exponentiate it, instead of computing a sigmoid and then taking a power:
\begin{eqnarray}
\log a_t = \log a^{c r_t} = \log \sigma(\Lambda) ^ {cr_t} = - c \softplus(\Lambda) \odot r_t.
\end{eqnarray}

 \paragraph{Gate behaviour}
 
Our gate is quite different than other standard gates in the literature.
In particular, most gating mechanisms, like the one used in Mamba and GRU, allow through the gate to interpolate fully between the hidden state and the new observation. 
Ours on the other hand is biased towards retaining information, and does not allow to fully discard the contribution of $h_{t-1}$ (this depends, however, on the value of $\Lambda$). 
To demonstrate this, we analyze the relative weight of $x_t$ compare to $h_{t-1}$ in the output $y_t$.
For a general recurrence we define this as:
\begin{eqnarray}
h_t = \alpha(r_t) h_{t-1} + \beta(r_t) x_t.
\end{eqnarray}
For our model we have $\alpha(r_t) = a_t = a^{c r_t}$ and $\beta(r_t) = \sqrt{1 - \alpha(r_t)^2}$.
For a standard GRU style gating we have $\alpha(r_t) = 1 - r_t$ and $\beta(r_t) = r_t$.
For Mamba, assuming in their notation $B = 1, C=1$, then $\alpha(r_t) = (1 - r_t)^{-A}$ and $\beta(r_t) = (1 - \alpha) / A$.
The behaviour of the different gating mechanisms is depicted in Figure~\ref{fig:different gates}, where for clarity we have also included the update value of the LRU \citep{orvieto2023resurrecting}, which has no gating. 
As can be seen, the Mamba gating is almost identical to the GRU for values of $A$ close to $1$, with minor deviations at smaller values.
On the other hand, our gating mechanism performs a very different non-linear interpolation between fully discarding the input $x_t$ and the update of the LRU.

\section{Complex-Gated Linear Recurrent Unit (CG-LRU) }
\label{appendix:complex_lru}

In Section~\ref{sec:RGLRU} we have defined our recurrent layer, however it can be further extended to use complex numbers. 
To achieve this we first parameterize a complex diagonal recurrence via $\tilde{a} = \sigma(\Lambda) e^{i\theta}$, where $\theta$ is a learnable parameter. 
In addition, we split the input $x_t$ along its channel dimensions, and interpret its first half as the real part of a complex vector, and the second part as the imaginary part of the same complex vector:
\begin{eqnarray}
x_t &= \begin{bmatrix} x_t^1 \\ x_t^2 \end{bmatrix} \\
\tilde{x}_t &= x_t^1 + i x_t^2.
\end{eqnarray}

With this we rewrite the equations for the LRU (see eq.~\ref{eq:RG_LRU}) as:
\begin{eqnarray}
 r_t &=& \sigma\left(W_a x_t + b_a\right), \quad \textcolor{gray}{\text{ \emph{recurrence gate}}} \\
 i_t & = & \sigma(W_x x_t + b_x), \quad \textcolor{gray}{\text{\emph{ input gate}}} \\
\tilde{a}_t &=& \tilde{a}^{c r_t}, \\
\tilde{h}_{t} &=& \tilde{a}_t \odot \tilde{h}_{t-1} + \sqrt{1-|\tilde{a}_t|^2} \odot (i_t \odot \tilde{x}_t).
\end{eqnarray}
We mark all complex variables with $\tilde{\cdot}$ for clarity.
Note that the number of dimensions of $r_t, i_t, \tilde{a}_t$ and $\tilde{h}_t$ are half of those of the real input $x_t$.
Finally, to compute the output we stack the real and imaginary part of $h_t$ into a single vector $y_t$:
\begin{eqnarray}
y_t &= \begin{bmatrix} \text{Real}(\tilde{h}_t) \\ \text{Imaginary}(\tilde{h}_t) \end{bmatrix}
\end{eqnarray}

\section{Model Scale Hyper-Parameters} \label{appendix:model_scale_hparams}
In Table \ref{tab:hypers}, we present the hyper-parameters of the models at different scales. These hyperparameters are shared for all the model families that we explored in this paper.

\begin{table}[htbp!]
\caption{Key model hyper-parameters considered for different model sizes. These hyperparameters are shared across different architectures we tested.}
\small
\centering
\begin{tabular}{lccccccl}
\toprule
\textbf{Model} & \textbf{Model Width} & \textbf{RNN Width} & \textbf{Depth} & \textbf{MLP Expansion} & \textbf{Attention} & \textbf{Training Tokens} \\
\textbf{Size} & $\bm{(D)}$ & $\bm{(D_{RNN})}$ & ($\bm{N}$) & \textbf{Factor ($\bm{M}$)} & \textbf{Heads}  & \textbf{(Optimal Scaling)} \\ \midrule
100M & 768 & 1024 & 12 & 3 & 6  & 1.9B \\
\rowcolor{tablerowcolor} 200M & 1024 & 1536 & 12 & 3 & 8  & 3.9B \\
400M & 1536 & 2048 & 12 & 3 & 12  & 7.8B \\
\rowcolor{tablerowcolor} 1.3B & 2048 & 2560 & 24 & 3 & 16  & 25B \\
3B & 3072 & 4096 & 24 & 3 & 24 & 60B \\
\rowcolor{tablerowcolor} 7B & 4096 & 5632 & 32 & 3 & 32  & 132.5B \\
14B & 5120 & 8192 & 40 & 3 & 40  & 300B
\end{tabular}
\label{tab:hypers}
\end{table}

\section{Efficient Linear Recurrences on Device}
\label{appendix:efficient_device}

\begin{table}[htbp!]
\centering
\begin{tabular}{l l} 
\toprule
\textbf{Specification} & \textbf{TPU-v3 Pod} \\
\midrule
HBM capacity & 32 GB \\
\rowcolor{tablerowcolor} HBM bandwidth & 900 GB/s \\
Peak MXU compute & 123 TFLOPs/s (bfloat16) \\ 
\rowcolor{tablerowcolor} Peak MXU FLOPs-to-byte-ratio & 136 \\
Peak VPU compute & 3.8 TFLOPs/s \\ 
\rowcolor{tablerowcolor} Peak VPU FLOPs-to-byte-ratio & 4.2 \\
\end{tabular}
\caption{Hardware specifications for a TPU-v3 pod.}
\label{appendix:tpu_specs}
\end{table}

\begin{figure}[t]
\centering
\subfigure[Scan runtimes]{\includegraphics{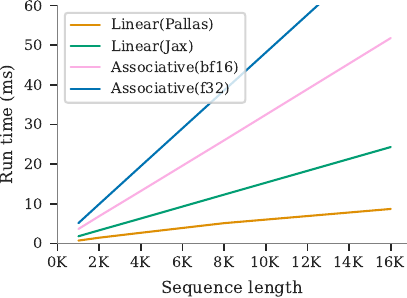}\label{fig:scan_a}}
\subfigure[
Hawk runtimes]{\includegraphics{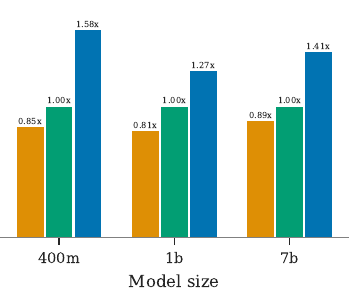}\label{fig:scan_b}}
\caption{a) Runtimes of different implementations of the scan operation on a TPU-v3 at different sequence lengths. The batch size of the input is fixed at 8 and the dimension of each token is 1024.
b) Relative runtimes of the Hawk model when using different implementations of the scan operation, in reference to the one with the native Jax scan implementation.}
\label{fig:scan}
\end{figure}

The initial step in computational optimization lies in identifying the primary performance bottleneck on the target hardware. 
For most accelerators, the key limiting factors are computational throughput (FLOPs/s) and memory bandwidth between the high-bandwidth memory (HBM) and the fast vector memory (VMEM). 
While factors like HBM capacity and host-device communication are relevant, techniques such as ZeRO sharding and pipelined data transfer offer practical mitigations. 
Modern accelerator designs often prioritize a high FLOPs-to-byte ratio to accommodate workloads where computations significantly outnumber memory transfers.  
We show the key specification of the TPU-v3 pod (two chips per pod) in Table~\ref{appendix:tpu_specs}, which we use for all our experiments.

\subsection{Matrix multiplication computation}
\label{appendix:matmul_example}

A typical matrix multiplication of a $D \times D$ matrix with a $D \times N$ matrix has $2ND^2$ FLOPs and $2(D^2 + 2ND)$ bytes to transfer (both read and write) which translates to $\frac{N D}{D + N}$ FLOPs/byte ratio.
When $D >> N$ and running on a TPU-v3 this implies that the dimension $N$ must be at least 136 to saturate the device, in which case the operation is ``compute bound'', or otherwise most of the time will be spent on waiting for memory transfers, in which case the operation is ``memory bound''. 

\subsection{Scan runtimes}
\label{sec:appendix_scan}

In Figure~\ref{fig:scan_a} we demonstrate that on a TPU-v3 our Pallas kernel achieves nearly x3 speed up compared to the naive Jax implementation.
In addition, the associative scan is significantly slower, even if fully run in bfloat16 precision.
Figure~\ref{fig:scan_b} demonstrates that these gains also translate to significant improvements of the overall training time per step of the full Hawk model even at the 7b scale.
For completeness we have also added the runtime of the associative scan, which can be up to 50\% slower.

\section{The Local Attention Window Size of Griffin}

\begin{figure}[t]
\centering
\includegraphics{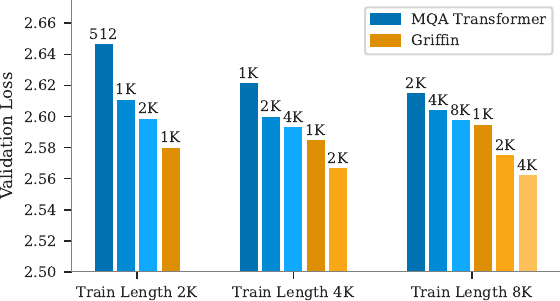}
\caption{Performance of 400M parameter Griffin and MQA Transformer models using different local attention window sizes and different training sequence lengths. The window sizes of the local attention layers are shown above each bar in the plot. We notice that a global attention MQA Transformer is much better than local attention variants of the MQA Transformer (where the window size is smaller than the training sequence length). Furthermore, we see that using a fixed local attention window size of 1024 (denoted `1K' in the plot) for the Griffin model outperforms all global attention and local attention MQA Transformer baselines across all training sequence lengths.}
\label{fig:LA-WS}
\end{figure}

Griffin uses both recurrent blocks as well as local attention layers in its temporal mixing blocks. For all experiments previously shown using a training sequence length of 2048, we use a local attention window size of 1024. We now investigate how the performance of different window sizes for the local attention layer varies with the training sequence length.

We consider 400M parameter models trained on sequence lengths of 2048, 4096 and 8192 tokens, where we keep the total number of training tokens fixed. For each sequence length, we train Griffin models using different local attention window sizes. As baselines, we train MQA Transformers using global attention layers, as well MQA Transformers using local attention layers with different window sizes. The results are shown in Figure~\ref{fig:LA-WS}, where the window sizes used are shown on top of each bar (MQA Transformer bars with window size equal to the training sequence length are the global attention MQA Transformer baseline).

From Figure~\ref{fig:LA-WS}, we see that remarkably, even when using a fixed window size of 1024 for the local attention layers in Griffin, it outperforms the global attention MQA Transformer baseline across all sequence lengths tested. However, it is worth noting that the performance gap between Griffin with local attention window 1024 and the global attention MQA Transformer reduces as the sequence length grows. Therefore, if the sequence length grows further, it is likely important to slowly also grow the local attention window size. In practice, the hardware used will also heavily determine the optimal local attention window size in terms of training and inference speed. Finally, we note that MQA Transformers purely using local attention (window sizes less than the training sequence length) perform significantly worse than both global attention MQA Transformers, as well as Griffin.

\section{Inference Speeds}

\subsection{Estimating memory-boundedness}\label{appendix:memory_bound}

The inference speed of language models at decode time is bounded by memory loading. As described already in \ref{sec:training_rnn} the linear RNN is memory bound. In the following we will show this is true for the other components (which are linear layers and self-attention) in our recurrent models and Transformer models.

\subsection{Estimating the memory boundedness of linear layers}
As shown in  \ref{appendix:matmul_example} the outer dimension (usually consisting of batch $B$ and sequence length $T$ dimensions) must be at least 136 in order to be compute bound. At decode time $T=1$ and if we assume $B \lesssim 128$ then any linear layers will be memory bound at decode time.

\subsection{Estimating the memory boundedness of self-attention}
In the following, we calculate the ratio of memory accesses to arithmetic operations for the attention computation for the $L$-th decode step, to show it is also memory-bound.

To simplify the following analysis, we assume that we start from an empty prompt (or equivalently assume that the  prefill contains 0 tokens).

When sampling auto-regressively from MHA or MQA, standard practice is to save the key and value vectors in a Key-Value (KV) cache. For $L$ tokens already sampled, the KV cache would therefore be of size $2 \times L \times H_k \times d_{head}$ for each sequence in the batch, where $H_k$ denotes the number of heads used for the keys and values, and $d_{head}$ denotes the dimension of the key and value vectors in each head.

For sampling the $L$-th token, once we calculate the query, key and value vectors corresponding to the $L$-th token. The attention weights and the output of the attention layer are then computed using the $L$-th key and value vectors in the KV cache. This requires $O(LD)$ operations overall and it requires loading the $O(L \times H_k \times d_{head})$ sized KV cache from HBM, for each sequence in the minibatch. The size of the KV cache, as well as the number of FLOPs, scales linearly with the batch size $B$.

For MHA, the number of heads for the key and values $H_k$ is typically equal to the number of heads used for the queries $H$. For MQA, a single head is used for keys and values, i.e., $H_k = 1$. Therefore for MQA, the size of the KV cache is a factor of $H_k$ smaller (i.e., of size $2 \times L \times d_{head}$). %

\begin{lstlisting}[language=Python]
def attention_sampling(q, k, v):
  """ Auto-regressive sampling via attention.
  For MHA, h_k = h. For MQA, h_k = 1.
  Args:
     q : The q vector for current token of shape [b, h, k]
     k : The keys of the current + previous tokens [b, L, h_k, k]
     v : the values of the current + previous tokens [b, L, h_k, v]
  """
  logits = einsum("bhk,bLk->bhL", q, k)  # O(bhLk)
  weights = softmax(logits)
  output = einsum("bhL,bLv->bhv", weights, v)  # O(bhLv)
  return output
\end{lstlisting}

For a batch size of $B$, the memory access to FLOPs ratio for the attention computation goes as $O(\frac{B \times L \times H_k \times d_{head}}{B \times L \times D})$. For typical Transformer architectures, $D =H \times d_{head}$ and further $H_k=H$ for MHA and $H_k=1$ for MQA.  Therefore the memory access to flops ratio is $O(1)$ for MHA and $O(1/H)$ for MQA. As explained in \ref{appendix:tpu_specs}, in order to be compute bound on TPU-v3 a FLOPs-to-byte ratio of 136 is required, and therefore both MHA and MQA would typically be memory bound. Nevertheless, MQA significantly speeds up Transformer inference (when compared to MHA), since it lowers the memory boundedness by a factor of $H$.

\subsection{Cache sizes}\label{appendix:cache_sizes}
In the following we do an analysis of the relative sizes of caches used in our recurrent and Transformers. All caches sizes scale linearly with batch size and in the following we assume $B=1$.
\subsubsection{The size of the KV cache}
For attention, the KV cache has size $2NTh_k d_{head}$, where $N$ denotes the number of attention layers (the depth), $T$ denotes the length of the sequence, $h_k$ denotes the number of KV heads and $d_{head}$ denotes the head dimension. Throughout this work, $d_{head} = 128$. For MHA, $h_k d_{head} = D$, while for MQA, $h_k = 1$. (We therefore expect MQA to be faster when decoding long sequences than MHA, since the size of the KV cache is significantly smaller and less memory needs to be moved.)

For either MHA or MQA the size of the KV cache can exceed the number of model parameters when the sequence length $T$ is large. We therefore expect to observe a transition from a `parameter bound' regime when the sequence length is short, during which the decoding speed is dominated by the time taken to load the model parameters on device, to a `cache bound' regime for large sequences, where the decoding speed is dominated by the time taken to load the KV cache.

\subsubsection{The size of the recurrent state}

The recurrent state of a single RG-LRU layer has size $D_{RNN}$, and the total state size for the entire Hawk model is $ND_{RNN} \approx 4BND/3$. Unlike the KV cache, this state does not grow with sequence length and is very small in comparison to parameter size. We therefore expect the decoding speed of RG-LRU to be dominated by the time taken to load the model parameters on device at all sequence lengths.

A similar consideration applies to the size of the 1D convolution state size. For a temporal filter width of size 4, the state has size $(4-1)D_{RNN}=3D_{RNN}=4D$ for each recurrent block which is also substantially smaller than parameter sizes.

\subsubsection{The local attention cache}

A single local MQA layer has cache size upper bounded by $2T_{WS}d_{head}$, where $T_{WS}$ denotes the local attention window size. So long as $T_{WS} \lesssim D^2/(Bd_{head})$, the size of the local attention cache is also small relative to the parameter count. We therefore expect the decoding speed of Griffin to be similar to the decoding speed of the Hawk model.

\section{Improving Next Token Prediction with Longer Contexts: Additional Results}\label{appendix:anthropic_plots}

\begin{figure}[ht]
\centering
\includegraphics{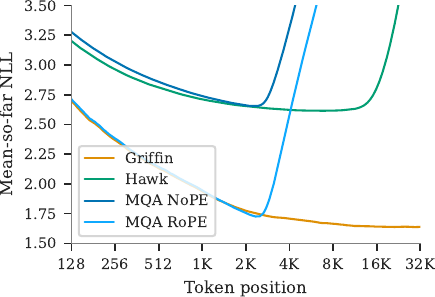}
\includegraphics{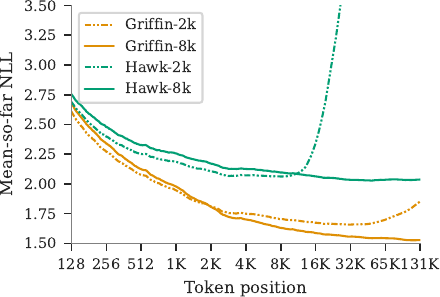}
\caption{
    The evaluation performance of 1B parameter models across a range of sequence lengths on held-out evaluation sets of ArXiv articles.
    On the left, we compare the performance of different models trained with sequence length 2048, evaluated with a sequence length of up to 32,768.
    On the right, we compare Griffin and Hawk when trained respectively on 2048 (2k) and 8192 (8k) sequence lengths.
    Results are qualitatively similar to the evaluation on Books presented in Figure~\ref{fig:longseq}.
}
\label{fig:longseq_arxiv}
\end{figure}

Figure \ref{fig:longseq_arxiv} shows an additional result demonstrating next token prediction performance at different context lengths on a held out dataset of arXiv articles. We find that the results on this dataset are qualitatively similar to the results shown in Figure \ref{fig:longseq}.

\section{Additional Details of the Copy and Retrieval Tasks}
\label{app:synthetic_tasks}

\begin{figure}[ht]
\centering
\subfigure[Selective Copying Task]{\includegraphics[width=0.4\linewidth]{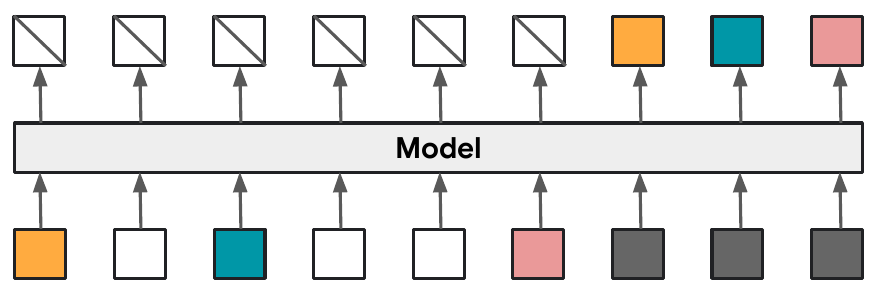}}
\hspace{1.25cm}
\subfigure[Induction Heads Task Task]{\includegraphics[width=0.4\linewidth]{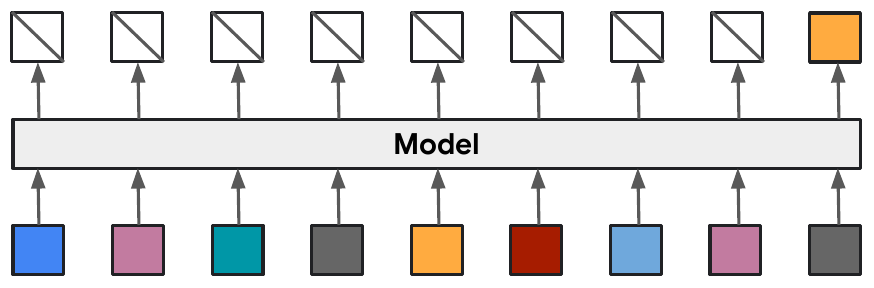}}
\caption{
    An illustration of the Selective Copying (left) and the Induction Heads tasks (right).
}
\label{fig:synthetic_tasks_illustration}
\end{figure}

Figure \ref{fig:synthetic_tasks_illustration} is an illustration of the Selective Copying and Induction Heads tasks.

In the Selective Copying task, the model needs to learn to copy data tokens (coloured tokens in Figure \ref{fig:synthetic_tasks_illustration}) from a sequence while ignoring noise tokens (white tokens in Figure \ref{fig:synthetic_tasks_illustration}). Crossed out tokens in the output in Figure \ref{fig:synthetic_tasks} denote tokens that are masked out in the loss.

In the Induction Heads task, the model needs to learn to recall the token immediately following a special token (black token in Figure \ref{fig:synthetic_tasks_illustration}). As before, crossed out tokens in the output denote tokens that are masked out in the loss.

\end{document}